\documentclass{article}
\PassOptionsToPackage{numbers, compress}{natbib}

 \usepackage[preprint]{neurips_2026}

\usepackage{wrapfig}
\usepackage[utf8]{inputenc} 
\usepackage[T1]{fontenc}    
\usepackage{hyperref}       
\usepackage{url}            
\usepackage{booktabs}       
\usepackage{amsfonts}       
\usepackage{nicefrac}       
\usepackage{microtype}      
\usepackage{xcolor}         

\usepackage{times}
\usepackage{amsmath}
\usepackage{graphicx}  
\usepackage{caption}   
\usepackage{multirow}
\usepackage{algpseudocode} 
\usepackage[ruled,vlined]{algorithm2e}
\usepackage{listings}
\usepackage{xcolor}
\usepackage{float}
\usepackage{enumitem}
\usepackage{bm}
\usepackage{amsthm}
\usepackage{amssymb}
\usepackage{subcaption}
\usepackage{wrapfig}
\usepackage{import}
\usepackage{tcolorbox}
\usepackage{todonotes}

\newtheorem{theorem}{Theorem}

\newtheorem{lemma}{Lemma}
\newtheorem{corollary}{Corollary}

\def\encs{u_\textrm{enc}}
\def\decs{u_\textrm{dec}}

\def\clayerFullName{Dual Reconstruction Smoothing}
\def\clayer{\textsf{\small DReS}}
\def\clayerB{\textsf{\small DReS}}
\def\clayerAbbv{\textsf{DReS}}

\title{{\clayerAbbv}: Dual Reconstruction Smoothing for Functional Regularization}

\author{%
  Parsa Moradi \\
  University of Minnesota\\
  \texttt{moradi@umn.edu} \\
   \And
  Tayyebeh Jahani-Nezhad \\
  Technical University Berlin \\
  \texttt{t.jahani.nezhad@tu-berlin.de} \\
  \And
  Hanzaleh Akbarinodehi \\
  University of Minnesota \\
  \texttt{akbar066@umn.edu} \\
  \And
  Mohammad Ali Maddah-Ali \\
  University of Minnesota \\
  \texttt{maddah@umn.edu} \\
}

\begin{document}

\maketitle
\begin{abstract}
Smoothness is a key inductive bias in machine learning and is closely related to generalization. Existing smoothness-inducing methods typically rely either on explicit gradient regularization, which often incurs substantial computational and memory overhead, or on data-mixing strategies, which are less naturally applicable to unsupervised and self-supervised settings. In this work, we propose 
\emph{\clayerFullName} ({\clayerB}), a nonparametric regularization framework that induces smoothness through a spline-based auxiliary branch with shared model parameters. The method introduces no additional trainable parameters and can be applied to arbitrary submodules, making it suitable for unsupervised, self-supervised, and supervised regimes. We show theoretically that the discrepancy between the target function and its {\clayerB} approximation is controlled by higher-order smoothness quantities of the function, establishing the method as an implicit higher-order smoothness regularizer. Empirically, {\clayerB} improves representation learning across several self-supervised methods, improves generation quality in generative modeling, and achieves strong performance relative to competitive baselines in supervised learning.
\end{abstract}

\section{Introduction}\label{sec:intro}
Generalization is a central objective in modern machine learning: models are
expected not only to fit training data, but also to perform reliably on unseen
examples and under distribution shifts or common corruptions
\citep{vapnik1998statistical,zhang2016understanding,hendrycks2019benchmarking,liu2021towards}.
A key inductive bias underlying such robustness is \emph{smoothness}: functions
whose outputs vary in a controlled manner with respect to their inputs tend to
generalize better and exhibit improved stability and robustness against perturbations
\citep{bousquet2002stability,sokolic2017robust,hein2017formal,miyato2018virtual,novak2018sensitivity, gulrajani2017improved}.


A large body of work has studied mechanisms for inducing smoothness in learned models. One prominent line of work enforces smoothness explicitly by penalizing derivatives of the model with respect to its input, thereby directly controlling local sensitivity to perturbations \citep{drucker1992improving,sokolic2017robust,ross2018improving,hoffman2019robust,finlay2021scaleable,foster2020improving}. However, these methods incur substantial computational and memory overhead, since optimizing input-gradient or Jacobian penalties typically requires differentiating through derivatives of the network. This substantially increases the size of the computational graph and may require second-order automatic differentiation, which is not always efficient or fully supported for all operations \citep{czarnecki2017sobolev,ross2018improving,etmann2019closer,finlay2021scaleable}.
Another widely used line of work promotes smoothness implicitly through data-centric strategies, including data augmentation and interpolation-based methods such as Mixup and its variants \citep{zhang2017mixup,verma2019manifold,berthelot2019mixmatch,yun2019cutmix,verma2022interpolation}. While effective, the smoothness bias induced by these methods is tied to the construction of synthetic examples. As a result, they often rely on label interpolation, pseudo-labels, or carefully chosen semantics-preserving transformations, which can limit their applicability when labels are unavailable, pseudo-labels are unreliable, or simple input-space transformations alter the underlying semantics \citep{zhang2017mixup,yun2019cutmix,berthelot2019mixmatch,sohn2020fixmatch,cubuk2019autoaugment}. Moreover, for Mixup-type methods, the induced smoothness typically takes the form of a strong affine interpolation bias, which can be restrictive.

In this work, rather than directly penalizing derivatives or modifying the training data, we introduce a new perspective on smoothness induction based on \emph{functional consistency between a model and its spline-based reconstruction along a smooth trajectory}. Given a model $f_\theta$ and an input batch, we construct an auxiliary predictor $\widetilde{f}_\theta$ using a pair of spline-based encoder and decoder operators. The encoder first fits a smooth spline trajectory through the input batch samples. Then a set of new samples is drawn from this \emph{input trajectory}. The model is evaluated on these new samples, after which the decoder fits a second smooth spline to the resulting model outputs, forming an \emph{output trajectory}. Sampling the output trajectory at specific points yields an approximation $\widetilde{f}_\theta$ of the model outputs on the input batch.
During training, the model is therefore
evaluated along two shared-parameter branches: a standard branch producing
$f_\theta(x)$, and an auxiliary branch producing $\widetilde{f}_\theta(x)$ through the
encoder--model--decoder pipeline (see Figure~\ref{fig:csm_training}).

Let $\mathcal{L}(\cdot;\mathcal{D})$ denote the task loss on the training set $\mathcal{D}$. We train the standard and auxiliary branches jointly by minimizing a weighted sum of the same task loss evaluated on their respective predictions:
\begin{align}
\mathcal{L}_{\mathrm{total}}(\theta)
=
(1-\mu)\,\mathcal{L}(f_\theta; \mathcal{D})
+
\mu\,\mathcal{L}(\widetilde{f}_\theta; \mathcal{D}),
\end{align}
where $\mu \in [0,1]$ controls the contribution of the auxiliary branch in the total loss. Because both predictors share parameters and are trained on the same objective, this formulation implicitly encourages agreement between the original model and its smoothed counterpart. Intuitively, the
encoder--decoder pathway suppresses high-frequency variation by forcing the
model to remain predictive after passing through a non-parametric smooth operator. This biases the learned model toward smoother
solutions without requiring explicit derivative penalties, higher-order
backpropagation, or additional trainable parameters.


We theoretically substantiate this intuition by showing that the proposed method is tied to higher-order smoothness of the learned model. In particular, we prove that the  discrepancy between applying
$f_\theta$ on the input trajectory and its spline-based surrogate $\widetilde{f}_\theta$ is controlled by higher-order derivative quantities.
These bounds provide a principled interpretation of the \clayerB{} as an implicit higher-order smoothness regularizer. 


Our method is entirely nonparametric and introduces no additional trainable parameters. The encoder--decoder operators act at the mini-batch level and incur only modest computational overhead. The framework is also highly flexible: it can be applied across a wide range of learning settings, and is particularly appealing in unsupervised and self-supervised regimes where many alternative smoothness-inducing methods are less naturally applicable.


We validate \clayerB{} across unsupervised, self-supervised, and supervised settings, demonstrating that it is not tied to a particular loss, architecture, or
supervision regime. In generative modeling, applying it to the generator of WGAN-GP~\citep{gulrajani2017improved} improves sample quality despite the critic-side gradient penalty in the baseline, indicating a complementary smoothness
effect on the generator mapping. In self-supervised learning, we incorporate it into the training of five well-established methods, BYOL~\citep{grill2020bootstrap}, MoCo v2~\citep{chen2020improved}, SimCLR~\citep{chen2020simple}, VICReg~\citep{bardes2021vicreg}, and Barlow Twins~\citep{zbontar2021barlow} on CIFAR-10 and CIFAR-100~\citep{krizhevsky2009learning} datasets.
Because these methods already rely on strong augmentation based invariances, the consistent improvements suggest that \clayerB{} contributes a complementary function-level smoothness bias. 
It improves representation quality under both linear probing and kNN evaluation, achieving up to $93.28\%$ and $70.60\%$ top-1 linear probing accuracy on CIFAR-10 and CIFAR-100, respectively, and yielding gains of up to $3.41\%$ in linear probing and $4.04\%$ in kNN accuracy across the evaluated methods and datasets. Taken together, these results support our central view of \clayerB{} as a
general-purpose smoothness-inducing module: it is label-free, nonparametric, and
modular, and can therefore be inserted into supervised, self-supervised, or
unsupervised pipelines without changing the underlying learning objective.

{\bf Contributions.}  
Our contributions are summarized as follows:
\begin{itemize}
    \item \textbf{Spline-based surrogate consistency.}
    We propose \emph{\clayerB}, a nonparametric regularization framework
    that induces smoothness by training a model together with a spline-smoothed
    auxiliary surrogate. The method introduces no additional trainable parameters,
    does not require explicit gradient penalties or label mixing, and can be
    applied to arbitrary network modules.

    \item \textbf{Higher-order smoothness regularization.}
    We provide a theoretical explanation of the regularization effect by relating
    the discrepancy between the model and its spline-smoothed surrogate to
    higher-order derivatives of the learned model. This establishes
    \clayerB{} as a higher-order smoothness regularizer without requiring higher-order backpropagation.

    \item \textbf{Broad validation across learning settings. }
    We demonstrate the effectiveness of \clayerB{} in generative modeling,
    self-supervised representation learning, and supervised classification. Across
    these settings, the method improves performance over the corresponding
    baselines and achieves competitive or superior results against established
    smoothness-inducing methods.
\end{itemize}

\begin{figure}[t]
  \centering
  \includegraphics[width=\textwidth]{figs/cs_framework_simple.png}
  \caption{The training pipeline of {\clayerB{}}. Input batch passes through both standard and auxiliary branches and the loss encourages both branches to perform well on the task.}
  \label{fig:csm_training}
\end{figure}

\section{Preliminaries}\label{sec:prelim}

In this section, we introduce notation, the block-level learning setup, and natural cubic spline interpolation, which is the basic building block of the proposed method.

{\bf Notations.} Throughout the paper, lowercase letters denote scalars (e.g., $x$), bold lowercase letters denote vectors (e.g., $\mathbf{x}$), and uppercase letters denote matrices (e.g., $A$). For a positive integer $n$, we define $[n]:=\{1,\dots,n\}$. Given a dataset $\mathcal{D}$, we denote by $f_\theta$ a model parameterized by $\theta$, and by $\mathcal{L}(f_\theta;\mathcal{D})$ the corresponding training loss. For a differentiable function $g:\mathbb{R}\to\mathbb{R}^d$, we write $g^{(k)}$ for its $k$-th derivative, with $g^{(0)}:=g$. The notation $\|\cdot\|_2$ denotes the Euclidean norm, and $\|g\|_{L^2(\Omega)}$ denotes the $L^2$ norm of a function $g$ over a domain $\Omega\subseteq\mathbb{R}$. The notation $a \asymp b$ means that there exist constants \(c_1,c_2>0\) such that $c_1 b \leq a \leq c_2 b$.

\subsection{Learning Setup}\label{sec:prob_formulation}

Let $\mathcal{D}=\{\xi_i\}_{i=1}^n$ denote a training set, where each $\xi_i$ is a task-dependent sample (e.g., $\xi_i=(\mathbf{x}_i,y_i)$ in supervised learning and $\xi_i=\mathbf{x}_i$ in unsupervised or self-supervised learning). We consider a network decomposed as
\(
\mathrm{net}_{\theta} = h_1 \circ f_{\theta} \circ h_2,
\)
where $h_1:\mathcal{V}\to\mathcal{Y}$ and $h_2:\mathcal{X}\to\mathcal{U}$ denote the parts of the network after and before the target block $f_{\theta}:\mathcal{U}\to\mathcal{V}$, respectively. The target block may represent either an intermediate submodule or the entire model, in which case $h_1(\cdot)$ and $h_2(\cdot)$ are identity mappings on their respective domains. Our goal is to regularize training so that the target block $f_\theta$ varies more smoothly over its input domain while preserving the original task objective.

Training is performed by minimizing a task-specific empirical loss $\mathcal{L}(\mathrm{net}_{\theta};\mathcal{D})$. For a mini-batch $\mathcal{B}=\{\mathbf{x}_i\}_{i\in[B]}$, we define the intermediate representations
\(
\mathbf{z}_i := h_2(\mathbf{x}_i)\in\mathcal{U}.
\)
The proposed method constructs, on each mini-batch, an auxiliary approximation of the target block evaluations $\{f_\theta(\mathbf{z}_i)\}_{i=1}^B$ and uses it to regularize training.

\subsection{Natural Cubic Splines}\label{sec:spline_prelim}

Let $\Omega\subseteq\mathbb{R}$ be an interval, and let
\(
t_1 < t_2 < \cdots < t_m
\)
be distinct knots in $\Omega$. Given observations $\{(t_i,v_i)\}_{i=1}^m$ with $v_i\in\mathbb{R}$, the natural cubic spline interpolant is the unique function $\widetilde g:\Omega\to\mathbb{R}$  solution of the variational problem
\begin{equation}\label{eq:natural_spline_var}
    \widetilde g
    =
    \operatorname*{argmin}_{g}
    \int_{\Omega} \bigl(g^{(2)}(t)\bigr)^2\,dt
    \quad
    \text{subject to }
    g(t_i)=v_i,\ \forall i\in[m].
\end{equation}
Thus, among all interpolating functions, the natural cubic spline is the one with minimum second-derivative energy. For fixed knots $\{t_i\}_{i=1}^m$, interpolation is linear in the values $\{v_i\}_{i=1}^m$~\citep{wahba1990spline}. Therefore, there exist basis functions $\{\phi_i\}_{i=1}^m$, depending only on the knots, such that
\(
    \widetilde g(t) = \sum_{i=1}^m v_i\,\phi_i(t)
\). For vector-valued data where $\mathbf{v}_i \in \mathbb{R}^d$, the spline is constructed componentwise, preserving this linearity across each dimension.

\section{\clayerFullName{} Module}\label{sec:cs_module}

The design of {\clayerB} is inspired by General Coded Computing (GCC), a framework for reliable distributed computation that is provably robust to straggling and adversarial servers~\citep{moradi2024coded,moradi2025general_adv,moradi2025general_prob}. Motivated by GCC, we propose the {\clayerB} module as a regularization mechanism for implicitly inducing functional smoothness during training. In this section, we first describe the architecture of the proposed module (Section~\ref{sec:cs_arch}) and its integration into the training pipeline (Section~\ref{sec:training_pipe}). We then explain how the resulting auxiliary branch promotes smoothness (Section~\ref{sec:highorder_smooth}) and discuss its computational complexity (Section~\ref{sec:comp_complexity}). Finally, we present the theoretical foundations of {\clayerB} by showing that the proposed training procedure acts as an implicit regularizer of higher-order derivatives of the learned model (Section~\ref{sec:theory}).

\begin{algorithm}[t]
  \caption{PyTorch-style pseudo-code for the {\clayerB} module}
  \label{alg:csm}
  \SetKwInput{Input}{Input}
  \SetKwInput{Output}{Output}
  \Input{Input batch $\mathbf{X}$ of shape $(B, d_{in})$; Model or submodule $f_\theta$.}
  \Output{Smoothed surrogate output $\widetilde{f}_\theta(\mathbf{X})$.}
  \vspace{0.5em}
\begin{lstlisting}[language=Python]
class SplineSmoothing(nn.Module):
    def __init__(self, B, N):
        super().__init__()
        self.set_points(B, N)
        
    def set_points(self, enc_points, dec_points):
        self.alpha = generate_encoding_points(enc_points) 
        self.beta = generate_decoding_points(dec_points) 
        self.enc_spline = NaturalCubicSpline(dim=enc_points) 
        self.dec_spline = NaturalCubicSpline(dim=dec_points)  
        
    def forward(self, X, f): 
        self.enc_spline.fit(self.alpha, X) # 1. Encoding
        x_coded = self.enc_spline.predict(self.beta)
        
        f_coded = f(x_coded) # 2. Computing
        
        self.dec_spline.fit(self.beta, f_coded) # 3. Decoding
        f_tilde = self.dec_spline.predict(self.alpha)
        return f_tilde
\end{lstlisting}
\end{algorithm}

\subsection{Architecture}\label{sec:cs_arch}

Let $\Omega:=(-1,1)$. The {\clayerB} module consists of three components:
(i) a spline-based encoder $\encs:\Omega\to\mathcal{U}$,
(ii) a computation function $f_{\theta}:\mathcal{U}\to\mathcal{V}$, and
(iii) a spline-based decoder $\decs:\Omega\to\mathcal{V}$.
Here, $\mathcal{U}$ and $\mathcal{V}$ denote the input and output domains of the target block $f_\theta$, and $\theta$ denotes its model parameters. Given a mini-batch of target-block inputs $\{\mathbf{z}_i\}_{i\in[B]} \subset \mathcal{U}$, the module produces auxiliary predictions $\{\widetilde f_\theta(\mathbf{z}_i)\}_{i\in[B]}$ through the following three-stage pipeline.

\begin{enumerate}[label={(\arabic*)}]
\setlength\itemsep{0.25em}

    \item \textbf{Encoding.}
    Let $\alpha_1<\cdots<\alpha_B$ be a fixed set of scalar \emph{encoding points} in $\Omega$, and let $\beta_1<\cdots<\beta_N$ be a fixed set of scalar \emph{decoding points} in $\Omega$, with $\{\alpha_i\}_{i\in[B]}\cap\{\beta_j\}_{j\in[N]}=\varnothing$. These points are chosen once for a given pair $(B,N)$ and reused across training batches; in our experiments, we use Chebyshev points (Section~\ref{sec:exp}). The natural cubic spline encoder $\encs$ is fitted to the pairs $\{(\alpha_i,\mathbf{z}_i)\}_{i=1}^B$, so that
    \begin{align}\label{eq:enc_approxm}
         \encs(\alpha_i)=\mathbf{z}_i, \qquad \forall i\in[B].
    \end{align}
    We then generate $N$ coded samples by evaluating the encoder at the decoding points,
    \(
        \widetilde{\mathbf{z}}_j=\encs(\beta_j)
    \) for $j\in[N]$. Since natural spline interpolation is linear in the values $\{\mathbf{z}_i\}_{i\in[B]}$, each coded sample $\widetilde{\mathbf{z}}_j$ can be interpreted as a linear combination of the original batch samples.

    \item \textbf{Computation.}
    The target block is evaluated on the coded samples, yielding the coded outputs
    \(
        \{f_{\theta}(\widetilde{\mathbf{z}}_j)\}_{j=1}^N.
    \)
    \item \textbf{Decoding.}
    The natural cubic spline decoder $\decs$ is fitted to the pairs
    $\{(\beta_j,f_{\theta}(\widetilde{\mathbf{z}}_j))\}_{j=1}^N$, so that
    \begin{align}\label{eq:dec_approxm}
       \decs(\beta_j)=f_{\theta}(\widetilde{\mathbf{z}}_j)
       =f_{\theta}(\encs(\beta_j)), \qquad \forall j\in[N].
    \end{align}
    Because the encoding and decoding points are disjoint, evaluating the decoder at the encoding points requires interpolation of the composed function $f_\theta\circ\encs$ from the values observed at $\{\beta_j\}_{j\in[N]}$. If the decoder well-approximates $f_\theta\circ\encs$ over $\Omega$, then
    \begin{align}\label{eq:rec_approxm}
        \decs(\alpha_i)\approx f_{\theta}(\encs(\alpha_i))
        =f_{\theta}(\mathbf{z}_i), \qquad \forall i\in[B].
    \end{align}

\end{enumerate}
     We define the auxiliary predictor by  
    $\widetilde f_{\theta}(z )\triangleq \decs(z)$. Algorithm~\ref{alg:csm} gives PyTorch-style pseudo-code for the module.

\subsection{Training Pipeline}\label{sec:training_pipe}

We now describe the integration of the {\clayerB} module into the training pipeline. Consider a neural network decomposed as
\(
    \mathrm{net}_{\theta}(\mathbf{x}) = 
    h_1\circ f_{\theta}\circ h_2(\mathbf{x}),
\)
where $h_1(\cdot)$ and $h_2(\cdot)$ denote the sub-networks following and preceding a target block $f_{\theta}(\cdot)$, respectively. Given an input batch $\{\mathbf{x}_i\}_{i=1}^B$, we first compute the intermediate representations $\mathbf{z}_i = h_2(\mathbf{x}_i)$ for $i\in[B]$. The \clayerB{} module is then applied to the target block $f_{\theta}$ using the batch $\{\mathbf{z}_i\}_{i=1}^B$ to generate the auxiliary predictions $\{\widetilde{f}_{\theta}(\mathbf{z}_i)\}_{i=1}^B$, following the encoding--computation--decoding pipeline detailed in Section~\ref{sec:cs_arch}. This construction induces two parallel branches during training. The \emph{standard branch} is defined as 
\(
    \mathrm{net}_{\theta}^{\mathrm{std}}(\mathbf{x}_i)
    =
    h_1(f_{\theta}(\mathbf{z}_i)),
\)
while the \emph{auxiliary branch} is defined as 
\(
    \mathrm{net}_{\theta}^{\mathrm{aux}}(\mathbf{x}_i)
    =
    h_1(\widetilde{f}_{\theta}(\mathbf{z}_i)).
\)
Both branches share the same model parameters $\theta$. In the auxiliary branch, the target block $f_{\theta}$ is replaced with its spline-based approximation. Let $\mathcal{L}_{\mathrm{std}}$ and $\mathcal{L}_{\mathrm{aux}}$ denote the task-specific losses associated with the standard and auxiliary branches, respectively. We define the
overall training objective as
\begin{equation}\label{eq:training_pipeline_loss}
    \mathcal{L}_{\mathrm{total}}
    =
    (1-\mu)\,\mathcal{L}_{\mathrm{std}}
    +
    \mu\,\mathcal{L}_{\mathrm{aux}},
\end{equation}
where $\mu\in[0,1]$ is a hyperparameter controlling the contribution of the auxiliary branch. 
Minimizing \eqref{eq:training_pipeline_loss} encourages the model $f_{\theta}$ to remain consistent with its spline-based surrogate $\widetilde{f}_{\theta}$ under the task objective. We will demonstrate that this consistency implicitly induces a functional smoothness regularization effect on the target block.

\subsection{Smoothness via Spline Reconstruction}
\label{sec:highorder_smooth}

We now provide intuition for why the objective in~\eqref{eq:training_pipeline_loss} promotes smoothness. This intuition is formally established later in Subsection~\ref{sec:theory}. The key observation is that, alongside the direct evaluation of the target block,
{\clayerB} computes an encoder--model--decoder composition.
In operator form, the auxiliary branch can be written as
\[
    \widetilde{f}_{\theta}
    =
    \mathcal{T}_{\beta\to\alpha}\circ f_{\theta}\circ \mathcal{T}_{\alpha\to\beta},
\]
where $\mathcal{T}_{\alpha\to\beta}$ denotes spline encoding from 
$\{(\alpha_i, \mathbf{z}_i)\}_{i\in[B]}$ to $\{(\beta_j, \Tilde{\mathbf{z}}_j)\}_{j\in[N]}$, 
and $\mathcal{T}_{\beta\to\alpha}$ denotes spline decoding $\{(\beta_j, f_\theta(\Tilde{\mathbf{z}}_j))\}_{j\in[N]}$ back to  $\{(\alpha_i, \tilde{f}_{\theta}(\mathbf{z}_i))\}_{i\in[B]}$. Thus, instead of using the values $\{f_\theta(\mathbf{z}_i)\}_{i\in[B]}$ directly, the auxiliary branch computes and uses $\{\tilde{f}(\mathbf{z}_i)\}_{i\in[B]}$.

This reconstruction is not arbitrary. By construction, the decoder $\decs$ is the natural cubic spline fitted to the coded outputs $\{(\beta_j, f_{\theta}(\widetilde{\mathbf{z}}_j))\}_{j\in[N]} = \{(\beta_j, (f_{\theta}\circ\encs)(\beta_j))\}_{j\in[N]}$, and is therefore the minimum-curvature interpolant of those values (see~\eqref{eq:natural_spline_var}). Consequently, the auxiliary branch favors functions for which the composition $f_\theta\circ\encs$ can be accurately recovered from the coded outputs by a low-curvature spline. Since the encoder $\encs$ is itself a natural cubic spline fitted to $\{(\alpha_i,\mathbf{z}_i)\}_{i\in[B]}$, this regularization acts along a spline-induced trajectory in the input space of the target block. Functions that vary too rapidly along the encoded trajectory $\encs(z)$ are reconstructed less accurately and are therefore disfavored by the auxiliary loss. More explicitly, the decoder admits the representation
\[
    \decs(z)
    =
    \sum_{j\in[N]}
    f_{\theta}(\widetilde{\mathbf{z}}_j)\,\phi_j(z),
    \qquad z\in\Omega,
\]
\begin{wrapfigure}{r}{0.49\textwidth} 
  \centering
  \vspace{-10pt} 
  \includegraphics[width=\linewidth]{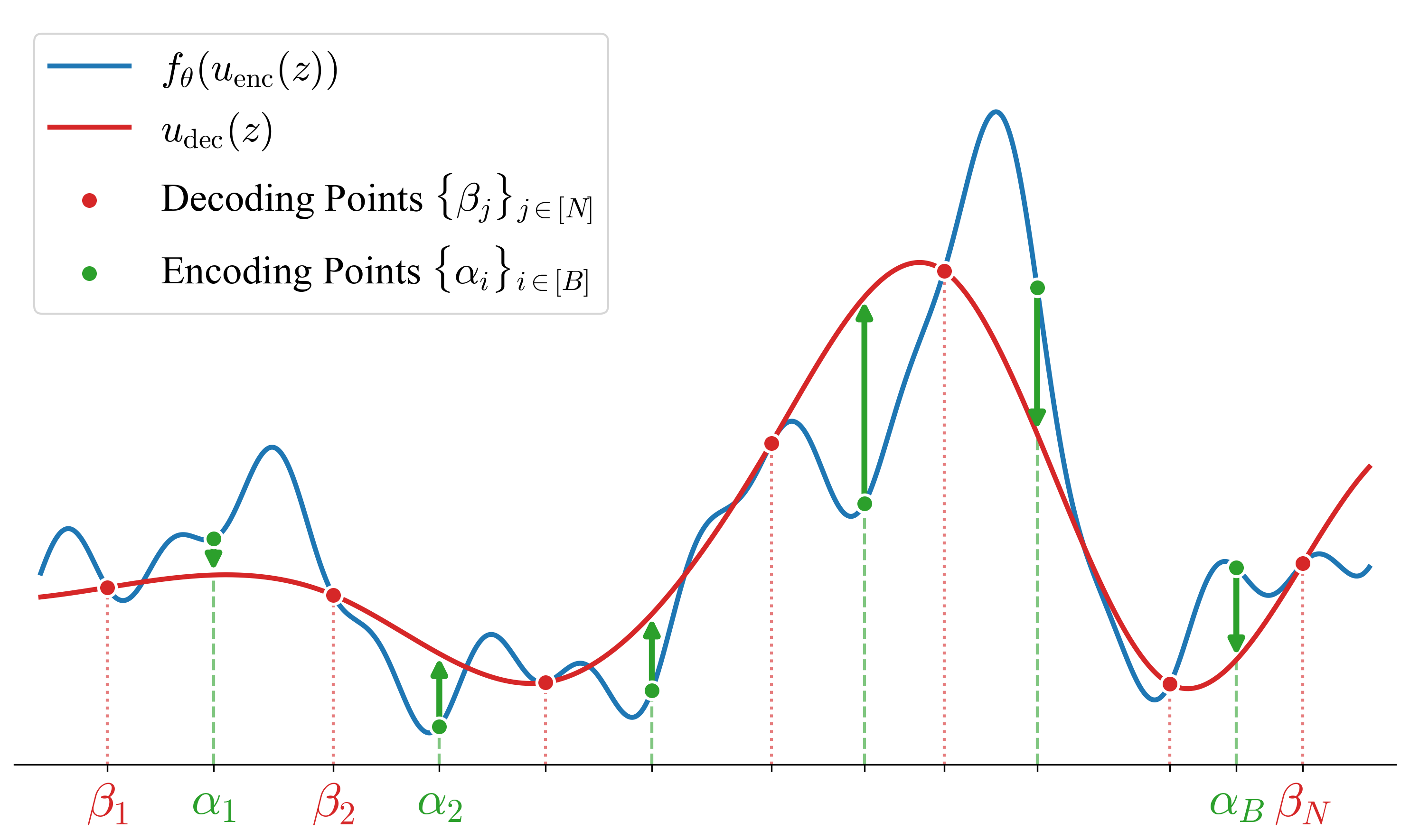}
  \caption{{\clayerB} promotes functional smoothness by pulling the target block along the spline-induced trajectory (blue curve) toward a low-curvature spline surrogate (red curve).}
  \label{fig:sm_csm}
\end{wrapfigure}
where $\phi_j$ denotes the natural spline basis function associated with the knot $\beta_j$. Hence, training with {\clayerB} encourages the reconstruction-consistency relation
\[
    f_{\theta}(\encs(z))
    \approx
    \sum_{j\in[N]}
    f_{\theta}(\widetilde{\mathbf{z}}_j)\,\phi_j(z),
    \qquad z\in\Omega.
\]
Since the right-hand side is a low-curvature spline reconstruction of the coded outputs, minimizing the auxiliary loss biases the target block toward smoother behavior along the spline-induced trajectory $\encs(t)$ (Figure~\ref{fig:sm_csm}). Thus, {\clayerB} acts as a flexible nonparametric smoothness regularizer that goes beyond explicit gradient penalties and pairwise interpolation, promoting functional smoothness through spline-based reconstruction.

\subsection{Computational Complexity}\label{sec:comp_complexity}

Consider training a neural network decomposed as
\(
\mathrm{net}_{\theta}(\mathbf{x}) = 
    h_1\circ f_{\theta}\circ h_2(\mathbf{x}),
\)
where \clayerB{} module is applied to $f_{\theta}$. Since both the encoder and decoder are nonparametric, {\clayerB} introduces no additional trainable parameters. The computational overhead of the proposed method comes from two distinct sources. The first is the forward-pass cost of the auxiliary branch.  In the auxiliary branch, the target block $f_\theta$ is evaluated on $N$ coded samples, while the subsequent block $h_1$ is evaluated on the reconstructed outputs $\{\widetilde{f}_\theta(\mathbf{z}_i)\}_{i\in[B]}$. Let $M$ denote the cost of a forward pass of a single sample through the full model, and let $M_f$ denote the cost of a forward pass through the target block alone. Then this computation cost of the auxiliary branch is $\mathcal{O}\!\left(
 MB + M_f(N-B)
    \right)$.

The second source of overhead comes from spline encoding and decoding. These include fitting the encoder and decoder and evaluating them at the decoding and encoding points, respectively. Since natural splines can be fitted and evaluated in linear time using a B-spline basis representation~\citep{de2001calculation,eilers1996flexible}, this overhead is small relative to the main training computation. More specifically, if the input and output dimensions of the target block are $d_{\mathrm{in}}$ and $d_{\mathrm{out}}$, respectively, then the computational complexities of the encoding and decoding steps are $\mathcal{O}((N+B)d_{\mathrm{in}})$ and $\mathcal{O}((N+B)d_{\mathrm{out}})$.

Thus the total forward-pass complexity of training with {\clayerB} is
\begin{equation}
    \mathcal{O}\!\left(
        MB + (N+B)(d_{\mathrm{in}}+d_{\mathrm{out}}) + MB + M_f(N-B)
    \right).
\end{equation}
When $N=\mathcal{O}(B)$ and $M, M_f \gg d_{\mathrm{in}}+d_{\mathrm{out}}$, the total cost remains of the same order as two standard forward passes, with only a small additional spline overhead.

To quantify the computational overhead of the proposed method, we compare the forward- and backward-pass runtimes of standard ERM, ERM + {\clayerB}, and ERM + Gradient Penalty (GP) training. As reported in Table~\ref{tab:infer} in Appendix~\ref{app:runtime}, explicit gradient regularization incurs a substantial cost during backpropagation. While its forward-pass runtime is comparable to that of ERM + {\clayerB}, its backward pass is more than an order of magnitude slower. By contrast, ERM + {\clayerB} introduces only moderate overhead relative to standard ERM and remains substantially more efficient than GP in the backward pass. Moreover, increasing the number of coded samples primarily affects the forward pass, without increasing the backward-pass cost. These results show that ERM + {\clayerB} offers a more efficient and scalable approach to imposing smoothness on the model than explicit gradient penalties.

\subsection{Theoretical Guarantees}\label{sec:theory}

In this section, we theoretically analyze the regularization effect of \clayerB{} by demonstrating that reducing discrepancy between the composed function $f\circ\encs$ and its {\clayerB} approximation $\widetilde f$
results in reducing norm of higher order derivatives of the model. 


Recall that $\{\alpha_i\}_{i=1}^{B},\{\beta_j\}_{j=1}^{N} \subset \Omega$ denote the encoding and decoding points, respectively. Recall that these sets are disjoint, i.e.,
\(
\{\alpha_i\}_{i=1}^{B} \cap \{\beta_j\}_{j=1}^{N} = \varnothing.
\)
We augment the decoding set with the boundary points by defining $\beta_0=-1$ and $\beta_{N+1}=1$, and form the sorted common refinement
\[
P := \{\alpha_i\}_{i=1}^{B} \cup \{\beta_j\}_{j=0}^{N+1}
= \{-1=\omega_0 < \omega_1 < \cdots < \omega_{N+B+1} = 1\}.
\]
This refinement induces a partition of $\Omega$ into subintervals $J_k := (\omega_k, \omega_{k+1})$ with lengths $\delta_k := \omega_{k+1} - \omega_k$ for $k \in \{0, \dots, B+N\}$. The next theorem provides both an upper and a lower bound on $\|f\circ\encs-\widetilde f\|_{L^2(\Omega)}^2$.

\begin{theorem}\label{thm:main_th}
Under the notation above, let $\Delta_{\beta} = \max_{j} (\beta_{j+1} - \beta_j)$ denote the maximum spacing between decoding points. Assume that on each refined interval $J_k$, the composition $f_\theta \circ \encs$ admits a fifth weak derivative satisfying $\|(f_\theta \circ \encs)^{(5)}\|_{L^2(J_k)} \le R$ for some constant $R > 0$ for all $k$'s. Then, there exist constants $C_1, C_2, C_3 > 0$ independent of $B$ and $N$ such that the {\clayerB} surrogate $\widetilde{f}_\theta$ satisfies:
\begin{align}
    \left\|f_\theta \circ \encs - \widetilde{f}_\theta \right\|_{L^2(\Omega)}^2
    &\ge
    C_1 \sum_{k=0}^{B+N}
    \min\!\left\{
        \frac{\left\|(f_\theta \circ \encs)^{(4)}\right\|_{L^2(J_k)}^8}{R^8}, \, 
        C_2\,\delta_k^8
    \right\}
    \left\|(f_\theta \circ \encs)^{(4)}\right\|_{L^2(J_k)}^2,
    \label{eq:lowerbound_main}
    \\
    \left\|f_\theta \circ \encs - \widetilde{f}_\theta \right\|_{L^2(\Omega)}^2
    &\le
    {C_3}\,\Delta^4_\beta \,
    \|(f_\theta \circ \encs)^{(2)}\|_{L^2(\Omega)}^2.
    \label{eq:upperbound_main}
\end{align}
\end{theorem}

The following corollary shows how the bounds in Theorem~\ref{thm:main_th} scale asymptotically as the number of coded samples increases.

\begin{corollary}[Asymptotic scaling]\label{cor:main_th}
Assume that the decoding points are quasi-uniform, satisfying $\underline{\Delta}/N \le \beta_{j+1} - \beta_j \le \overline{\Delta}/N$ for $j\in \{0,\dots,N+1\}$ and some constants $\underline{\Delta}, \overline{\Delta} > 0$. Then, for sufficiently large $N$, there exist constants $C_4, C_5 > 0$ independent of $B$ and $N$ such that:
\begin{align}\label{eq:asymp}
    \frac{C_4}{N^8}
    \left\|(f_\theta \circ \encs)^{(4)}\right\|_{L^2(\Omega)}^2
    \le
    \left\|f_\theta \circ \encs - \widetilde{f}_\theta \right\|_{L^2(\Omega)}^2
    \le
    \frac{C_5}{N^4}
    \left\|(f_\theta \circ \encs)^{(2)}\right\|_{L^2(\Omega)}^2.
\end{align}
\end{corollary}

A detailed proof of Theorem~\ref{thm:main_th} and Corollary~\ref{cor:main_th} are provided in Appendix~\ref{sec:appendix_proofs}. The theorem yields a two-sided characterization of the discrepancy $\|f\circ\encs-\widetilde f\|_{L^2(\Omega)}$. The upper bound shows that when $f\circ\encs$ has small second-order variation, the decoder yields a high-fidelity approximation. Conversely, the lower bound shows that a small discrepancy necessarily requires control of a fourth-order smoothness quantity of $f\circ\encs$. Moreover, by the chain rule, the derivatives $(f\circ\encs)^{(2)}$ and $(f\circ\encs)^{(4)}$ depend on derivatives of the model $f$ up to order two and four, respectively. In this sense, reducing the discrepancy between $f\circ\encs$ and $\widetilde f$ implicitly constrains higher-order variation of the learned model, up to fourth order, along the spline-induced trajectory $\encs(\cdot)$.

Taken together, Theorem~\ref{thm:main_th} and Corollary~\ref{cor:main_th} clarify the role of the proposed auxiliary branch. They show that the approximation error between $f_\theta\circ\encs$ and its {\clayerB} surrogate is closely tied to higher-order smoothness of the composed function and the number of coded samples.

\section{Experiments}\label{sec:exp}

In this section, we evaluate the effectiveness of incorporating {\clayerB} into training across multiple learning settings and evaluation metrics. A key advantage of {\clayerB} is its flexibility: it can be applied to arbitrary submodules of a network and does not require label information. This makes it naturally compatible with a broad range of learning paradigms. We begin with self-supervised learning, where we incorporate the proposed method into the training of several self-supervised representation learning methods (Section~\ref{sec:selfsup}). We then study generative modeling as an example of unsupervised learning (Section~\ref{sec:gen}). Finally, we evaluate {\clayerB} in supervised learning and compare it against well-established baselines across multiple benchmarks (Section~\ref{sec:exp_sup}). All experiments are implemented in PyTorch~\citep{paszke2019pytorch} and conducted on a single machine with an NVIDIA RTX 6000 GPU.

In all experiments, following~\citep{jahani2022berrut,moradi2024coded}, we use first-kind Chebyshev points for encoding and second-kind Chebyshev points for decoding, namely
\(
\alpha_i = \cos(-\frac{(2i-1)\pi}{2B}),
\beta_j = \cos(-\frac{(j-1)\pi}{N}),
\)
for $i\in[B]$ and $j\in[N]$. This choice is motivated both by their strong empirical performance~\citep{jahani2022berrut} and by their favorable approximation-theoretic properties~\citep{phillips2003interpolation,trefethen2019approximation}.

\begin{table}[t]
\centering
\caption{Linear probing and kNN test accuracy (\%) on CIFAR-10 and CIFAR-100 for self-supervised representation learning methods with a ResNet-18 backbone. We compare each baseline with its regularized counterpart. {\clayerB} consistently improves representation quality across all methods.}
\label{tab:selfsup}
\renewcommand{\arraystretch}{1}
\setlength{\tabcolsep}{4.9pt}
\begin{tabular}{lcccc|cccc}
\toprule
 & \multicolumn{4}{c}{CIFAR-10} & \multicolumn{4}{c}{CIFAR-100} \\
\cmidrule(lr){2-5} \cmidrule(lr){6-9}
 & \multicolumn{2}{c}{Linear Prob} & \multicolumn{2}{c}{kNN} & \multicolumn{2}{c}{Linear Prob} & \multicolumn{2}{c}{kNN} \\
\cmidrule(lr){2-3} \cmidrule(lr){4-5} \cmidrule(lr){6-7} \cmidrule(lr){8-9}
 Method & Original & + \clayerAbbv & Original & + \clayerAbbv & Original & + \clayerAbbv & Original & + \clayerAbbv \\
\midrule
BYOL & 91.68 & {\bf 92.74}& 90.07 & {\bf 90.76}& 64.67 & {\bf 68.08}& 58.09 & {\bf 62.13}\\
SimCLR& 90.16 & {\bf 91.17}& 89.02 & {\bf 90.02} & 65.61 & {\bf 66.62} & 62.50 & {\bf 63.02}\\
MoCo v2 & 92.57 & {\bf 93.28}& 91.79 & {\bf 92.69} & {69.40} & {\bf 70.60} & 66.10 & {\bf 68.69} \\
VICReg & 90.10 & {\bf 90.50}& 88.63 & {\bf 88.86} & 62.20 & {\bf  64.91} & 57.90 & {\bf 59.81}\\
BT & 89.50 & {\bf 89.90}& 88.12 & {\bf 88.68} & 65.35 & {\bf 66.03 } & 60.37 & {\bf 61.65} \\
\bottomrule
\end{tabular}
\end{table}
\begin{table}[t]
\centering
\caption{Comparison of IS and FID for generated images on CIFAR-10 and CelebA.}
\label{tab:unsup}
\renewcommand{\arraystretch}{1}
\setlength{\tabcolsep}{8pt}
\begin{tabular}{lccc}
\toprule
\multirow{2}{*}{Method} 
& \multicolumn{2}{c}{CIFAR-10} 
& CelebA \\
\cmidrule(lr){2-3} \cmidrule(lr){4-4}
& IS $\uparrow$ & FID $\downarrow$ & FID $\downarrow$ \\
\midrule
WGAN-GP 
& $7.08 \pm 0.07$ 
& $26.93 \pm 0.61$ 
& $28.22 \pm 0.17$ \\

WGAN-GP + \clayerAbbv 
& $\mathbf{7.38 \pm 0.06}$ 
& $26.94 \pm 0.89$ 
& $\mathbf{24.58 \pm 0.62}$ \\
\bottomrule
\end{tabular}
\end{table}

\subsection{Self-Supervised Learning}\label{sec:selfsup}

We exploit the flexibility of {\clayerB} to regularize the training of several widely used self-supervised representation learning methods, with the goal of learning smoother and therefore more transferable representations. Specifically, we consider SimCLR~\citep{liu2020simple}, MoCo v2~\citep{chen2020improved}, BYOL~\citep{grill2020bootstrap}, VICReg~\citep{bardes2021vicreg}, and Barlow Twins (BT)~\citep{zbontar2021barlow} with ResNet18\citep{he2016deep} encoder backbone on the unlabeled CIFAR-10 and CIFAR-100~\citep{krizhevsky2009learning} datasets. For each method, we first train the original baseline and then train its {\clayerB}-regularized counterpart using the same experimental setup and hyperparameters. 

To evaluate representation quality, we freeze the learned encoder and assess it using both linear probing and $k$-nearest-neighbor (kNN) classification on the test set. Table~\ref{tab:selfsup} compares the original methods with their {\clayerB}-regularized versions. As shown in the table, incorporating {\clayerB} consistently improves performance across all considered self-supervised methods on both CIFAR-10 and CIFAR-100. Full implementation details and architectural designs are provided in Appendix~\ref{app:selfsup_detail} and Figure~\ref{fig:app_ssl_csm_all}.

\subsection{Generative Modeling}\label{sec:gen}
We next evaluate the effectiveness of {\clayerB} in unsupervised generative modeling. Specifically, we incorporate {\clayerB} into the training of Wasserstein GAN with Gradient Penalty (WGAN-GP)~\citep{arjovsky2017wasserstein,gulrajani2017improved}. Although WGAN-GP already includes an explicit gradient penalty to promote smoothness, we show that augmenting its training with {\clayerB} yields further improvements in generation quality.

Prior work has shown that regularizing the discriminator can improve GAN training stability and performance~\citep{zhang2017mixup,verma2019manifold}. In contrast, because mixup-style methods rely on label information, they cannot be directly applied to the generator. Here, we use the {\clayerB} module to regularize the \emph{generator} of WGAN-GP. Implementation details are provided in Appendix~\ref{app:unsup_detail}. Table~\ref{tab:unsup} reports the Fréchet Inception Distance (FID)~\citep{heusel2017gans} and Inception Score (IS)~\citep{salimans2016improved} on CIFAR-10 and CelebA~\citep{liu2018large}, two standard metrics for generative quality. As shown in the results, regularizing the generator with {\clayerB} improves IS on CIFAR-10 and substantially improves FID on CelebA, indicating improved generation quality overall.

\subsection{Supervised Learning}\label{sec:exp_sup}
We now evaluate the effectiveness of the {\clayerB} module in supervised learning. In particular, we compare its test performance against standard empirical risk minimization (ERM) and two widely used smoothness-inducing baselines: mixup~\citep{zhang2017mixup} and manifold mixup~\citep{verma2019manifold}.

To evaluate the method across diverse datasets and model families, we conduct experiments on CIFAR-10, CIFAR-100~\citep{krizhevsky2009learning}, and TinyImageNet~\citep{le2015tiny}. For CIFAR-10, we use PreActResNet18~\citep{he2016identity}; for CIFAR-100, we use WideResNet28-10~\citep{zagoruyko2016wide}; and for TinyImageNet we use ResNet50~\citep{goyal2017accurate}. 

In all supervised experiments, we find that the best performance is obtained when the {\clayerB} module is applied to the full network (see Appendix~\ref{app:abl} for a complete ablation study). Table~\ref{tab:sup_gen} in Appendix~\ref{app:sup_detail} summarizes the test results across datasets and architectures.  {\clayerB} outperforms both ERM and mixup-based baselines in nearly all benchmarks. Additional experiments and hyperparameter selection details are provided in Appendix~\ref{app:sup_detail}.

\section{Related Work}\label{sec:rel_work}

{\bf Coded Computing.}
The design of {\clayerB} is inspired by coded computing, a framework originally developed for reliable distributed computation in the presence of stragglers and adversarial workers~\citep{lee2017speeding,yu2019lagrange}. The central idea is to inject redundancy into the computation process through encoding and decoding, so that the desired output can still be recovered even when some workers fail or return corrupted results. Early work in coded computing focused on linear, bilinear, and polynomial computations, where algebraic structure makes exact recovery tractable~\citep{lee2017speeding,yu2017polynomial,yu2019lagrange,lee2017high,jahani2018codedsketch,codedpri}. More recent work has extended these ideas beyond polynomial settings by using tools from approximation and learning theory to design coding schemes for general functions~\citep{jahani2022berrut,moradi2024coded,moradi2025general_adv,moradi2025general_prob}. Our work is most closely related to this latter line, particularly General Coded Computing~\citep{moradi2024coded}. Unlike prior methods, however, our goal is not reliable distributed execution. Instead, we repurpose the encoding--decoding structure as a training-time regularization mechanism for inducing smoothness in learned models.

{\bf Gradient Regularization.} A classical approach to inducing smoothness is to penalize input derivatives. This idea dates back to double backpropagation~\citep{drucker1992improving}, where the norm of the input gradient is explicitly regularized during training. Variants of this principle appear in adversarial robustness, generative modeling, and representation learning via gradient penalties, Jacobian regularization, or derivative-based objectives~\citep{gulrajani2017improved,ross2018improving,jakubovitz2018improving,hoffman2019robust,finlay2020train}. These methods provide a direct way to control local sensitivity, but they typically require higher-order differentiation, increased memory usage, and additional computational overhead. In contrast, our method does not directly penalize derivatives. Instead, it induces smoothness implicitly through an auxiliary spline-based branch optimized under the original task loss. Thus, it offers a scalable approach with minimal computational overhead and no additional trainable parameters.


 {\bf Data-Centric Methods.} Another broad strategy improves generalization by inducing smoothness through data augmentation and interpolation, most notably Mixup \citep{zhang2017mixup}. By training on convex combinations of inputs and labels, Mixup encourages linear behavior between examples, which can be interpreted as a form of data-adaptive regularization of the model’s first- and second-order derivatives \citep{zhang2020does}. This paradigm has been extended to hidden representations \citep{verma2019manifold}, region-level mixing \citep{yun2019cutmix}, and semi-supervised consistency objectives \citep{verma2022interpolation}. While successful, these methods primarily require label information and promote first-order smoothness along pairwise directions. Our approach differs in two critical respects: first, it naturally extends to unsupervised and self-supervised settings because it does not rely on label mixing. Second, rather than enforcing consistency between predictions on convex combinations of inputs and the corresponding convex combinations of labels, DReS encourages consistency between the model and its spline-smoothed reconstruction along smooth trajectories.

\section{Conclusion}

We presented {\clayerB}, a spline-based auxiliary module for inducing smoothness during training. The proposed framework constructs a nonparametric auxiliary branch around a target block and optimizes it under the same task objective as the standard branch, without explicit derivative penalties or additional trainable parameters.

Our theoretical analysis shows that the discrepancy between the target block and its {\clayerB} approximation is controlled by second- and fourth-order smoothness quantities, providing a principled justification for interpreting the method as an implicit higher-order smoothness regularizer. More broadly, {\clayerB} provides a unified and practical framework for smoothness induction across learning paradigms, especially in self-supervised and unsupervised regimes where alternative approaches are often less natural or less effective. These results show that {\clayerB} is a simple, flexible, and effective approach for improving generalization.

\newpage
\appendix
\section{Appendix}

\section{Experimental Details}\label{app:exp_details}
In this section, we provide full experimental details for all settings. Training with {\clayerB} introduces two main hyperparameters: the loss weight $\mu$ and the number of coded samples $N$. These parameters play complementary roles: $\mu$ governs the strength of the auxiliary regularization, while $N$ determines the resolution of the spline-based surrogate. Empirically, {\clayerB} is fairly robust to the choice of $\mu$, and $\mu=0.5$ gives the best performance in most settings (see Appendix~\ref{app:mu} for a comparison across different values of $\mu$).

At the beginning of training, the target block is typically far from smooth, so the auxiliary branch provides only a coarse approximation. For this reason, we start with a small auxiliary weight and gradually increase $\mu$ using a sigmoid schedule, allowing the main task loss to dominate the early optimization stage.

We also schedule the number of coded samples $N$. When $N$ is fixed, the auxiliary branch has finite approximation resolution. Early in training this is not the dominant limitation, since the discrepancy between the standard and auxiliary branches is mainly caused by the nonsmoothness of the target block. Later in training, however, as the target block becomes smoother, the remaining discrepancy is increasingly governed by the resolution of the spline surrogate. At that stage, increasing $N$ refines the surrogate and allows the auxiliary branch to remain informative, thereby providing a extra push toward smoother behavior (see Figure~\ref{fig:sup_err}).

\subsection{Self-Supervised Learning Details}\label{app:selfsup_detail}

We evaluate {\clayerB} on five widely used self-supervised representation learning methods: Barlow Twins, BYOL, MoCo v2, SimCLR, and VICReg. All experiments are conducted for $1,000$ epochs using a ResNet-18~\citep{he2016deep} backbone on both the CIFAR-10 and CIFAR-100 datasets. The self-supervised methods are implemented using the Lightly~\citep{susmelj2020lightly} library for SSL models and PyTorch Lightning~\citep{falcon2019pytorch}. Unless otherwise specified, we use cosine annealing learning-rate scheduling with warm-up. The baseline hyperparameters mostly follow the settings recommended in the original papers and implemented in the solo-learn~\citep{da2022solo} repository. For each method, the {\clayerB}-regularized experiment is identical to its corresponding baseline in all training hyperparameters and implementation details, except for the addition of the auxiliary branch and the associated {\clayerB}-specific parameters, such as $\mu$ and the number of coded samples $N$.

\begin{figure*}[t]
    \centering

    \begin{subfigure}[t]{0.44\textwidth}
        \centering
        \includegraphics[width=\linewidth]{figs/simclr.png}
        \caption{{\clayerB} integrated into SimCLR.}
        \label{fig:app_simclr_csm}
    \end{subfigure}
    \hfill
    \begin{subfigure}[t]{0.55\textwidth}
        \centering
        \includegraphics[width=\linewidth]{figs/byol.png}
        \caption{{\clayerB} integrated into BYOL.}
        \label{fig:app_byol_csm}
    \end{subfigure}

    \vspace{0.8em}

    \begin{subfigure}[t]{0.54\textwidth}
        \centering
        \includegraphics[width=\linewidth]{figs/bt_vic.png}
        \caption{{\clayerB} integrated into VICReg / Barlow Twins.}
        \label{fig:app_vicreg_bt_csm}
    \end{subfigure}
    \hfill
    \begin{subfigure}[t]{0.45\textwidth}
        \centering
        \includegraphics[width=\linewidth]{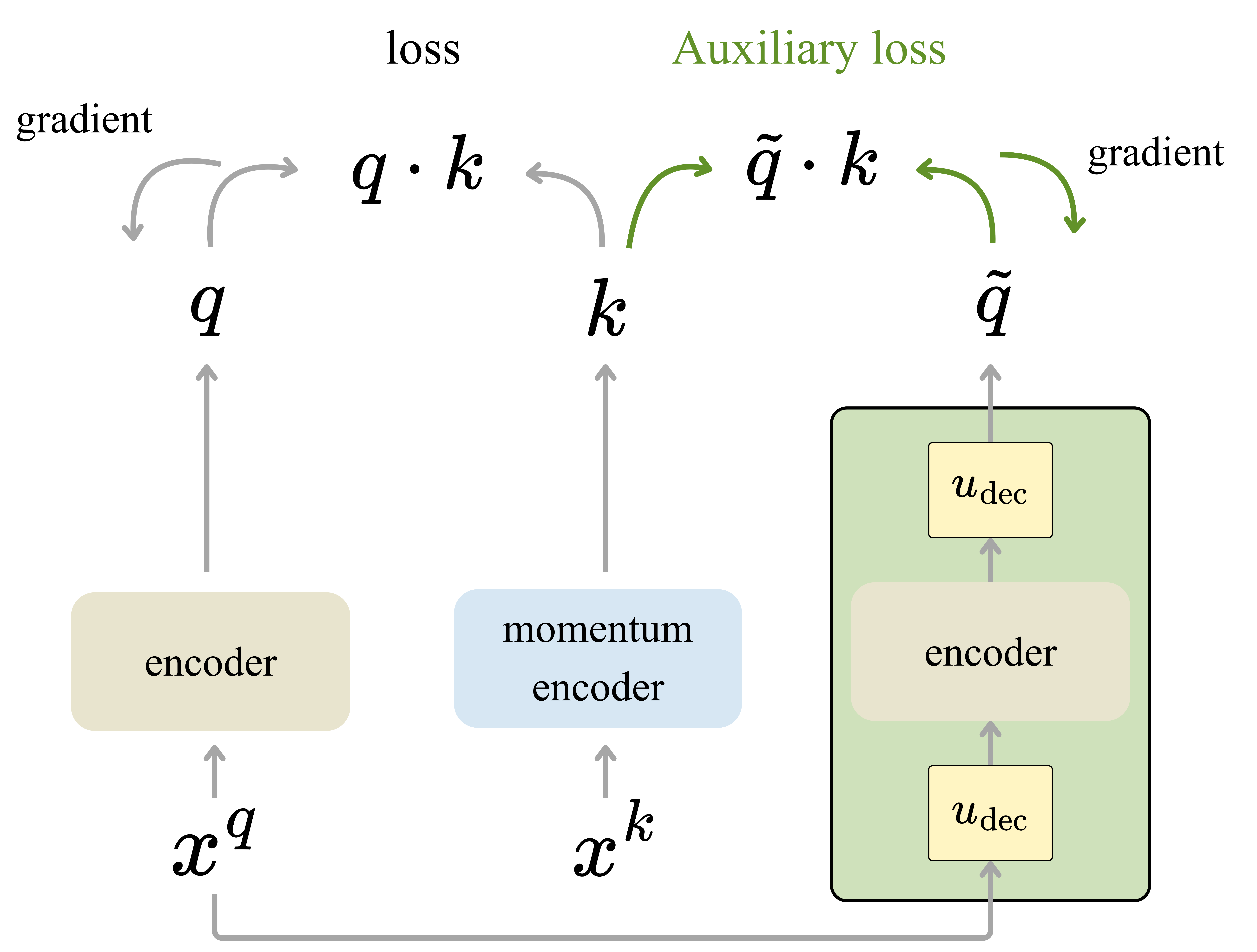}
        \caption{{\clayerB} integrated into MoCo v2.}
        \label{fig:app_moco_csm}
    \end{subfigure}

    \caption{Architectural integration of {\clayerB} into different self-supervised learning frameworks. Each panel follows the notation of the corresponding original method.}
    \label{fig:app_ssl_csm_all}
\end{figure*}

{\bf Barlow Twins.}
For Barlow Twins, we use the LARS optimizer with learning rate $0.4$, batch size $512$, and $1000$ training epochs. The learning rate is scheduled using cosine annealing with $10$ warm-up epochs. The projector hidden dimension is set to $4096$, the projector output dimension to $256$, and the predictor hidden dimension to $4096$. We apply {\clayerB} to the full network, including both the backbone and the projection head. We set the initial number of coded samples to $N=B=512$, use $\mu=0.3$, and increase $N$ to $1.5B$ after $500$ epochs (see Figure~\ref{fig:app_vicreg_bt_csm}).

{\bf BYOL.}
For BYOL, we use batch size $512$, $1000$ epochs, and the LARS optimizer with initial learning rate $0.4$ and weight decay $10^{-5}$. The projector hidden dimension is $4096$, the projector output dimension is $256$, and the predictor hidden dimension is $4096$. We apply {\clayerB} only to the online branch, i.e., the branch through which gradients are backpropagated, and not to the momentum target branch. We initialize the number of coded samples at $N=B$, set $\mu=0.5$, and increase $N$ to $1.5B$ after $500$ epochs (Figure~\ref{fig:app_byol_csm}).

{\bf MoCo v2.}
For MoCo v2, we use batch size $256$, memory bank size $32{,}768$, and SGD with momentum. The initial learning rate is $0.3$, the weight decay is $10^{-4}$, the projector hidden dimension is $2048$, and the projector output dimension is $256$. The contrastive loss temperature is set to $0.2$. We apply {\clayerB} to the online encoder only, and not to the momentum encoder. We set $\mu=0.5$, initialize the coded sample size at $N=312$, and increase it to $1.5\times 312$ after $500$ epochs (Figure~\ref{fig:app_moco_csm}).

{\bf SimCLR.}
For SimCLR, we use batch size $512$, initial learning rate $0.6$, the LARS optimizer, and weight decay $10^{-4}$. The projector hidden dimension is $2048$, the projector output dimension is $256$, and the contrastive loss temperature is $0.2$. We apply {\clayerB} to the full encoder branch. We set $\mu=0.3$, initialize the coded sample size at $N=1.5B$, and increase it to $2B$ after $500$ epochs (Figure~\ref{fig:app_simclr_csm}).

\paragraph{VICReg.}
For VICReg, we use batch size $256$, initial learning rate $0.3$, and weight decay $10^{-4}$. The projector hidden dimension is $2048$ and the projector output dimension is $2048$. We apply {\clayerB} to the full network. We set $\mu=0.5$, initialize the coded sample size at $N=256$, and increase it to $1.5\times 256$ after $800$ epochs (see Figure~\ref{fig:app_vicreg_bt_csm}).

\paragraph{Downstream evaluation.}
After training each self-supervised model, we extract the backbone encoder and freeze all of its parameters. We then evaluate the learned representations using both linear probing and $k$-nearest-neighbor (kNN) classification on the corresponding downstream task, i.e., CIFAR-10 or CIFAR-100.

For linear probing, we attach a single linear classifier on top of the frozen encoder and train only this classifier using the labeled training set. We use cosine annealing learning-rate scheduling, and the initial learning rate is selected separately for each method and dataset by sweeping over a set of candidate values and choosing the one that gives the best validation performance. After selecting the best learning rate, we retrain the linear classifier with that setting and report the final accuracy on the test set.

For kNN evaluation, we similarly tune the number of neighbors $k$ for each method and dataset using validation performance, and then report the resulting test accuracy with the best selected value of $k$.

\subsection{Unsupervised Learning Details}\label{app:unsup_detail}

For the unsupervised experiments with WGAN-GP, training is run for $100{,}000$ iterations with batch size $64$. We set the number of coded samples to $N=96$ and use $\mu=0.5$. The initial learning rate is $2\times10^{-4}$, the critic is updated five times per generator step, and the gradient penalty coefficient is set to $\lambda_{\mathrm{gp}}=10$. Optimization is performed with Adam using betas $(0.0,0.9)$.

\begin{figure}[t]
    \centering
    \begin{subfigure}[b]{0.48\linewidth}
        \centering
        \includegraphics[width=\linewidth]{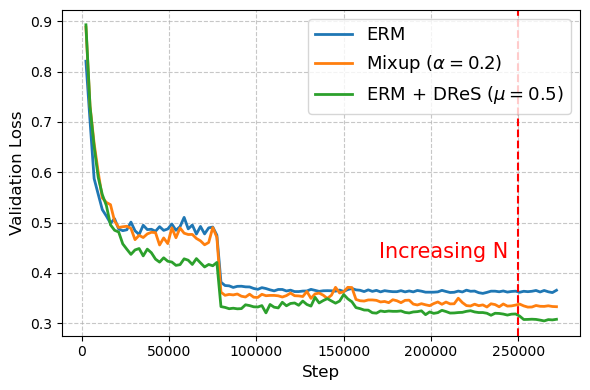}
        \caption{Validation Loss (TinyImageNet)}
        \label{fig:sup_err}
    \end{subfigure}
    \hfill
    \begin{subfigure}[b]{0.48\linewidth}
        \centering
        \includegraphics[width=\linewidth]{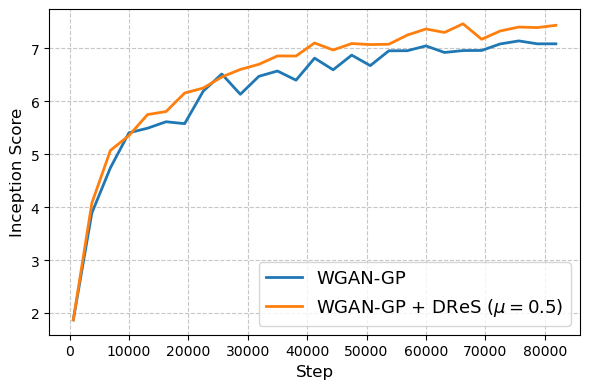}
        \caption{Inception Score (CIFAR-10)}
        \label{fig:unsup_is}
    \end{subfigure}
    
    \caption{\textbf{Performance dynamics during training.} \textbf{(a)} Validation loss trajectories on TinyImageNet. The {\clayerB} method exhibits a more stable descent and achieves lower validation error compared to standard training and Mixup. \textbf{(b)} Comparison of the Inception Score (IS) during the training of a WGAN-GP on CIFAR-10. Regularizing the generator using the {\clayer} module improves the IS.}
    \label{fig:perf_plot}
\end{figure}

For Inception Score (IS), we use $50{,}000$ generated samples. For FID, we follow the standard protocol and compare the statistics of $50{,}000$ generated images with those of the real dataset. Figure~\ref{fig:unsup_is} plots the IS curves during training for WGAN-GP and WGAN-GP + {\clayerB}.

\subsection{Supervised Learning Details}\label{app:sup_detail}

For all supervised experiments, we train models on CIFAR-10, CIFAR-100, and TinyImageNet using a common optimization setup. Each experiment is repeated over five independent runs with different random seeds, and we report the mean and standard deviation.

For CIFAR-10 and CIFAR-100, training is performed for $350$ epochs with an initial learning rate of $0.1$, decayed by a factor of $10$ at epochs $100$ and $250$.

For Tiny ImageNet, we use the same training length of $350$ epochs and the same initial learning rate of $0.1$, with decay by a factor of $10$ at epochs $100$, $200$, and $300$.


Across all experiments, we use stochastic gradient descent (SGD) with momentum $0.9$ and batch size $128$.

\renewcommand{\arraystretch}{1.2}
\begin{table}[H]
\centering
\captionof{table}{Comparisons of accuracies (\%) on test data.}
\label{tab:sup_gen}
\begin{tabular}{c|ccc}
\hline
 & CIFAR-10 & CIFAR-100 & TinyImageNet \\
 Method & PARN18 & WRN28-10 & RN50 \\
\hline
ERM              & $93.8\pm 0.2$ & $76.7\pm 0.3$ & $62.9\pm 0.9$ \\
Mixup            & $95.6\pm 0.2$ & $79.8\pm 0.4$ & $65.4\pm 1.0$\\
Manifold Mixup & $95.43\pm 0.12$  & \bm{$81.1 \pm 0.4$} & \bm{$67.4\pm 0.3$}\\
\hline
ERM + \clayerAbbv   & \bm{$95.8\pm 0.1$} & $79.9\pm 0.4$ & \bm{$67.1\pm 0.5$} \\
\hline
\end{tabular}
\end{table}

\subsubsection{Decision Boundary Comparison}

\begin{figure}[t]
    \centering   \includegraphics[width=1.0\linewidth]{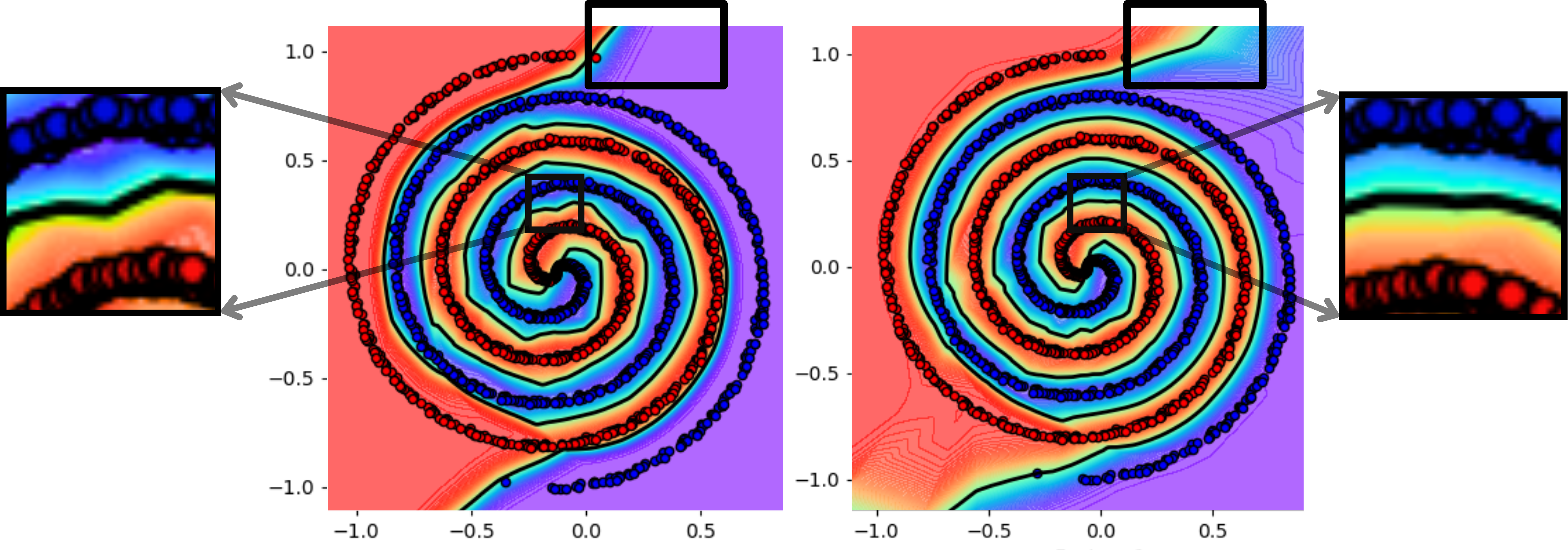}
    \caption{Decision boundaries of a 3-layer MLP on the spiral dataset trained with Mixup (left) and {\clayerB} (right). {\clayerB} yields visibly smoother and more stable decision boundaries. In particular, the level-set trajectories in the right panel are more widely and uniformly spaced than those in the left panel (see, for example, the upper-right corners), indicating more gradual variation of the learned model and hence a smoother decision surface.}
\label{fig:mixup_relation}
\end{figure}

To qualitatively compare the smoothness induced by {\clayerB} and Mixup, we train a 3-layer MLP with hidden widths $[1000,100,10]$ on the 2D spiral dataset and visualize both the decision boundaries and level sets of the learned classifier (see Figure~\ref{fig:mixup_relation}).

For Mixup, the interpolation coefficient is sampled from $\mathrm{Beta}(\alpha,\alpha)$, where $\alpha$ is selected based on validation performance. For {\clayerB}, we set $\mu=0.3$, which gives the best validation performance.

As shown in Fig.~\ref{fig:mixup_relation}, training with {\clayerB} produces visibly smoother and more stable decision boundaries than Mixup. This is also reflected in the level sets: in the {\clayerB} panel, neighboring contours are more widely and uniformly spaced, indicating more gradual variation of the learned model and hence a smoother decision surface.

\subsubsection{Hyperparameter Selection}\label{app:hyper}

{\bf Baselines.} For Mixup and Manifold Mixup, the mixing coefficient $\lambda$ is sampled from a Beta distribution $\mathrm{Beta}(\alpha,\alpha)$. We follow the optimal configurations reported in the original works~\cite{zhang2017mixup,verma2019manifold}: for Mixup, we set $\alpha = 1.0$ for CIFAR-10/100 and $\alpha = 0.2$ for TinyImageNet. For Manifold Mixup, we select the eligibility set of intermediate layers as described in \citet{verma2019manifold}, with $\alpha = 2.0$ for CIFAR-10/100 and $\alpha = 0.2$ for TinyImageNet.

{\bf {\clayerB} Parameters.} Empirically, we find that initializing with $N=B$ (the batch size) and $\mu=0.5$ provides an optimal trade-off between regularization strength and predictive accuracy. Additionally, we increase the number of coded samples $N$ to $N=1.5\times B$ after epoch $300$.

\section{Proofs of Main Results}
\label{sec:appendix_proofs}

\begin{proof}[Proof of Theorem \ref{thm:main_th}]
We divide the proof into two parts: deriving the lower bound via local Gagliardo--Nirenberg interpolation~\citep{leoni2024first} on the common refinement, and establishing the global upper bound using standard spline approximation theory. Let $e := f_\theta \circ \encs - \widetilde{f}_\theta$ denote the interpolation error.

\paragraph{Part 1: The Lower Bound \eqref{eq:lowerbound_main}.}
By construction, the sorted common refinement $P = \{\omega_k\}_{k=0}^{B+N+1}$ contains all encoding points of the encoder $\encs$ and all decoding points of the decoder $\widetilde{f}_\theta$. Therefore, strictly within the interior of any subinterval $J_k = (\omega_k, \omega_{k+1})$, neither $\encs$ nor $\widetilde{f}_\theta$ possess any knots. 

Since $\widetilde{f}_\theta$ is a cubic spline, its restriction to $J_k$ is a pure cubic polynomial, meaning its higher-order derivatives vanish identically: $\widetilde{f}_\theta^{(4)} = \widetilde{f}_\theta^{(5)} = 0$ on $J_k$. Consequently, for almost every $x \in J_k$, the derivatives of the error simplify exactly to the derivatives of the target composition:
\begin{equation}
    e^{(4)} = (f_\theta \circ \encs)^{(4)} \quad \text{and} \quad e^{(5)} = (f_\theta \circ \encs)^{(5)}.
\end{equation}

We apply the Gagliardo--Nirenberg interpolation inequality locally on $J_k$~\citep[Theorem~13.61]{leoni2024first}. For the $L^2$ norms with derivative orders $4$ and $5$, and local length scale $\delta_k$, there exist absolute constants $c_1, c_2 > 0$ such that:
\begin{equation}
    \left\|e^{(4)}\right\|_{L^2(J_k)} \le c_1 \left\|e^{(5)}\right\|_{L^2(J_k)}^{4/5} \|e\|_{L^2(J_k)}^{1/5} + \frac{c_2}{\delta_k^4} \|e\|_{L^2(J_k)}.
\end{equation}
Substituting the local derivative equalities yields:
\begin{equation}
    \left\|(f_\theta \circ \encs)^{(4)}\right\|_{L^2(J_k)} \le c_1 \left\|(f_\theta \circ \encs)^{(5)}\right\|_{L^2(J_k)}^{4/5} \|e\|_{L^2(J_k)}^{1/5} + \frac{c_2}{\delta_k^4} \|e\|_{L^2(J_k)}.
\end{equation}
For the sum on the right-hand side to bound the left-hand side, at least one of the two terms must be greater than or equal to $\frac{1}{2} \left\|(f_\theta \circ \encs)^{(4)}\right\|_{L^2(J_k)}$. This creates two possible cases for the lower bound of $\|e\|_{L^2(J_k)}$:

\noindent \textit{Case 1: The fractional term dominates.} 
\begin{equation}
    c_1 \left\|(f_\theta \circ \encs)^{(5)}\right\|_{L^2(J_k)}^{4/5} \|e\|_{L^2(J_k)}^{1/5} \ge \frac{1}{2} \left\|(f_\theta \circ \encs)^{(4)}\right\|_{L^2(J_k)}.
\end{equation}
Solving for $\|e\|_{L^2(J_k)}$ and squaring both sides gives:
\begin{equation}
    \|e\|_{L^2(J_k)}^2 \ge \frac{1}{1024\,c_1^{10}} \frac{\left\|(f_\theta \circ \encs)^{(4)}\right\|_{L^2(J_k)}^{10}}{\left\|(f_\theta \circ \encs)^{(5)}\right\|_{L^2(J_k)}^8}.
\end{equation}
Using the assumption that $\left\|(f_\theta \circ \encs)^{(5)}\right\|_{L^2(J_k)} \le R$, we obtain:
\begin{equation}
    \|e\|_{L^2(J_k)}^2 \ge c_1' \frac{\left\|(f_\theta \circ \encs)^{(4)}\right\|_{L^2(J_k)}^8}{R^8} \left\|(f_\theta \circ \encs)^{(4)}\right\|_{L^2(J_k)}^2,
\end{equation}
where $c_1' = 1 / (1024\,c_1^{10})$.

\noindent \textit{Case 2: The linear term dominates.}
\begin{equation}
    \frac{c_2}{\delta_k^4} \|e\|_{L^2(J_k)} \ge \frac{1}{2} \left\|(f_\theta \circ \encs)^{(4)}\right\|_{L^2(J_k)} \implies \|e\|_{L^2(J_k)}^2 \ge \frac{\delta_k^8}{4\,c_2^2} \left\|(f_\theta \circ \encs)^{(4)}\right\|_{L^2(J_k)}^2.
\end{equation}
Let $c_2' = 1 / (4\,c_2^2)$.

Since the true local error squared is bounded from below by the minimum of Case 1 and Case 2, we can factor out $\left\|(f_\theta \circ \encs)^{(4)}\right\|_{L^2(J_k)}^2$ to write:
\begin{equation}
    \|e\|_{L^2(J_k)}^2 \ge \min\!\left\{ c_1' \frac{\left\|(f_\theta \circ \encs)^{(4)}\right\|_{L^2(J_k)}^8}{R^8}, \, c_2' \delta_k^8 \right\} \left\|(f_\theta \circ \encs)^{(4)}\right\|_{L^2(J_k)}^2.
\end{equation}
By defining the global constants $C_1 = c_1'$ and $C_2 = c_2' / c_1'$, the local bound scales as desired. Because the refined intervals $J_k$ partition $\Omega$ up to a set of measure zero, the global $L^2$ error over $\Omega$ is strictly the sum of the squared local errors:
\begin{equation}
    \|e\|_{L^2(\Omega)}^2 = \sum_{k=0}^{B+N} \|e\|_{L^2(J_k)}^2.
\end{equation}
Substituting the local minimum bound into the summation yields the lower bound \eqref{eq:lowerbound_main}.

\paragraph{Part 2: The Upper Bound \eqref{eq:upperbound_main}.}
To establish the upper bound, we leverage classical approximation results for cubic spline interpolation~\citep{ragozin1983error, utreras1988convergence}. 

Let $\mathcal{H}^2(\Omega)$ denote the Sobolev space of functions $g: \Omega \to \mathbb{R}^d$ such that $g$ and its first two weak derivatives are square-integrable. By construction, the encoder $\encs$ is a natural cubic spline, and since the model $f_\theta$ is assumed to be sufficiently smooth (admitting up to a fifth weak derivative), the composition $g = f_\theta \circ \encs$ belongs to $\mathcal{H}^2(\Omega)$. 

The following lemma characterizes the approximation error for such functions:

\begin{lemma}[{\cite[Theorem 4.10]{ragozin1983error}}]\label{lemma:spline_noiseless_app}
Let $g \in \mathcal{H}^{2}(\Omega)$ be a function defined on the interval $\Omega=(a,b)$, and let $\{y_i\}_{i=1}^n$ be noiseless observations satisfying $y_i=g(t_i)$ at points
\(
a<t_1<\cdots<t_n<b.
\)
Let $\widetilde{g}$ denote the natural cubic spline interpolant of the data $\{(t_i, y_i)\}_{i\in[n]}$. Define the discretization factor by
\begin{align}
    L = D \cdot \Delta_{\max}^{4},
\end{align}
where $\Delta_{\max} = \max_i(t_{i+1}-t_i)$ and $D$ is a constant. Then, there exists a constant $G > 0$ such that:
\begin{align}\label{eq:noiseless_bound_app}
    \| g - \widetilde{g} \|_{L^2(\Omega)}^2 \le G \cdot L \cdot \|g^{(2)}\|_{L^2(\Omega)}^2.
\end{align}
\end{lemma}

We apply Lemma~\ref{lemma:spline_noiseless_app} by identifying $g := f_\theta \circ \encs$ and $\widetilde{g} := \widetilde{f}_\theta$. In our framework, the interpolation points are the decoding points $\{\beta_j\}_{j=1}^N$, and the maximum interval length is denoted by $\Delta_\beta = \Delta_{\max}$. Therefore, we have:
\begin{align}
    \left\|f_\theta \circ \encs - \widetilde{f}_\theta \right\|_{L^2(\Omega)}^2 \le G D \cdot \Delta_\beta^4 \cdot \left\|(f_\theta \circ \encs)^{(2)}\right\|_{L^2(\Omega)}^2.
\end{align}
By defining the constant $C_3 :=GD$, we obtain the desired upper bound:
\begin{align}
    \left\|f_\theta \circ \encs - \widetilde{f}_\theta \right\|_{L^2(\Omega)}^2 \le C_3 \Delta_\beta^4 \left\|(f_\theta \circ \encs)^{(2)}\right\|_{L^2(\Omega)}^2,
\end{align}
which completes the proof of Theorem~\ref{thm:main_th}.
\end{proof}

\begin{proof}[Proof of Corollary \ref{cor:main_th}]
We first address the lower bound. In the common refinement $P$, let us identify the "clean" subintervals. Because there are exactly $B$ interior encoding knots $\alpha_i$, at most $B$ of the original $N+1$ decoding intervals $(\beta_j, \beta_{j+1})$ are split. Let $\mathcal{I}_{\text{clean}}$ denote the index set of the refined intervals $J_k$ that are identical to unsplit decoding intervals. There are at least $N + 1 - B$ such clean intervals, and by the quasi-uniformity assumption, their lengths satisfy $\delta_k \ge \underline{\Delta}/N$ for all $k \in \mathcal{I}_{\text{clean}}$.

Since all terms in the summation of \eqref{eq:lowerbound_main} are non-negative, the global error squared is bounded below by the sum over just the clean intervals:
\begin{equation}
    \left\|f_\theta \circ \encs - \widetilde{f}_\theta \right\|_{L^2(\Omega)}^2 \ge C_1 \sum_{k \in \mathcal{I}_{\text{clean}}} \min\!\left\{ \frac{\left\|(f_\theta \circ \encs)^{(4)}\right\|_{L^2(J_k)}^8}{R^8}, \, C_2\,\delta_k^8 \right\} \left\|(f_\theta \circ \encs)^{(4)}\right\|_{L^2(J_k)}^2.
\end{equation}

For $k \in \mathcal{I}_{\text{clean}}$, as $N \to \infty$, the length $\delta_k \asymp N^{-1} \to 0$. The fractional term inside the minimum scales as $\delta_k^4 \asymp N^{-4}$, while the second term scales as $\delta_k^8 \asymp N^{-8}$. Therefore, for sufficiently large $N$, the $\mathcal{O}(N^{-8})$ linear term strictly dominates the minimum. Therefore,  for sufficiently large $N$ we obtain
\begin{align}
    \left\|f_\theta \circ \encs - \widetilde{f}_\theta \right\|_{L^2(\Omega)}^2 &\ge C_1 \sum_{k \in \mathcal{I}_{\text{clean}}} C_2 \left(\frac{\underline{\Delta}
    }{N}\right)^8 \left\|(f_\theta \circ \encs)^{(4)}\right\|_{L^2(J_k)}^2 \nonumber \\
    &= \frac{C_1 C_2 \underline{\Delta}^8}{N^8} \left\|(f_\theta \circ \encs)^{(4)}\right\|_{L^2\left(\bigcup_{\text{clean}} J_k\right)}^2.
\end{align}

The total length of the excluded "split" intervals is at most $B \max_{j}(\beta_{j+1} - \beta_j) \le B\overline{\Delta}/N$. Since $B$ is fixed, this measure vanishes as $N \to \infty$. Thus, the integral over the clean intervals converges to the global integral over $\Omega$. For sufficiently large $N$, we have:
\begin{equation}
    \left\|(f_\theta \circ \encs)^{(4)}\right\|_{L^2\left(\bigcup_{\text{clean}} J_k\right)}^2 \ge \frac{1}{2} \left\|(f_\theta \circ \encs)^{(4)}\right\|_{L^2(\Omega)}^2.
\end{equation}
Defining $C_4 = \frac{1}{2} C_1 C_2 c^8$ yields the left-hand side of \eqref{eq:asymp}.

For the upper bound, we evaluate \eqref{eq:upperbound_main} under the assumption $\Delta_\beta \le \overline{\Delta}/N$ which yields $\Delta^4_\beta \le \overline{\Delta}\,^4/N^4$. Absorbing $C^4$ into the absolute constant $C_5$, we directly obtain the right-hand side of \eqref{eq:asymp}:
\begin{equation}
    \left\|f_\theta \circ \encs - \widetilde{f}_\theta \right\|_{L^2(\Omega)}^2 \le \frac{C_3}{N^4} \left\|(f_\theta \circ \encs)^{(2)}\right\|_{L^2(\Omega)}^2,
\end{equation}
which completes the proof.
\end{proof}

\section{Ablation Study}\label{app:abl}

\subsection{Runtime Analysis}\label{app:runtime}
\begin{table}[H]
\centering
\captionof{table}{Forward- and backward-pass runtimes for standard ERM training, ERM + {\clayerB}, and ERM + Gradient Penalty (GP) using ResNet50 on CIFAR-100. While {\clayerB} introduces only moderate overhead, gradient penalty incurs a much larger backward-pass cost.}
\centering
\renewcommand{\arraystretch}{1} 
\begin{tabular}{lll}
\toprule
Experiment & Forward Pass Time (s) & Backward Pass Time (s)  \\
\midrule
ERM          & $0.0259 \pm 0.0027$ & $0.0059	\pm 0.0011$  \\
ERM + \clayerB{} (N=128) & $0.0518 \pm 0.0047$  & $0.0132 \pm 0.0032$ \\
ERM + \clayerB{} (N=190) & $0.0695 \pm	0.0059 $  & $0.0131 \pm	0.0020$ \\
ERM + Gradient Penalty            & $0.0531 \pm	0.0042$  & $0.1580 \pm	0.0006$ \\
\bottomrule
\end{tabular}
\label{tab:infer}
\end{table}

\subsection{Effect of number of coded samples ($N$)}\label{app:n}
\begin{table}[H]
\centering
\captionof{table}{Test accuracy (\%) of training with {\clayerB} module for different number of coded samples ($N$) on CIFAR-10 with batch size $128$.}
\label{tab:eff_N}
\begin{tabular}{l|c}
\hline
 $N$ & Acc  \\
\hline
$110$              & $95.6$  \\
$128$              & $95.8$  \\
$135$          & $95.0$  \\
$150$          & $94.8$  \\
$170$          & $94.5$  \\
$190$          & $94.2$  \\
\hline
\end{tabular}
\end{table}

\subsection{Effect of $\mu$}\label{app:mu}
\begin{table}[H]
\centering
\captionof{table}{Test accuracy (\%) of training with {\clayerB} module for different $\mu$ on CIFAR-10.}
\label{tab:eff_mu}
\begin{tabular}{l|c}
\hline
 $\mu$ & Acc  \\
\hline
$0.1$              & $95.1$  \\
$0.2$          & $95.4$  \\
$0.4$          & $95.7$  \\
$0.5$          & $95.9$  \\
$0.6$          & $95.8$  \\
$0.8$          & $95.4$  \\
$1.0$          & $95.0$  \\
\hline
\end{tabular}
\end{table}

\subsection{Effect of the layer set}\label{app:layer}

We conducted additional experiments in which \clayerB{} was selectively applied to different layers or blocks within the ResNet architecture. Specifically, we applied the \clayerB{} module to various subsets of ResNet blocks in the PreActResNet-18 model (with total 4 ResNet blocks) for the CIFAR-10 task. The results are shown in Table~\ref{tab:layer}.

\begin{table}[h]
\centering
\captionof{table}{Effect of coded-path block selection on test accuracy and loss.}
\begin{tabular}{c|cc}
\toprule
\textbf{Set of blocks in the coded path} & \textbf{Test Acc} & \textbf{Test Loss} \\
\midrule
0           & 94.6  & 0.25  \\
3           & 94.6  & 0.26  \\
0,1         & 94.8  & 0.23  \\
0,2         & 94.8  & 0.23  \\
1,3         & 94.8  & 0.235 \\
0,1,2       & 94.5  & 0.24  \\
0,1,2,3     & 95.1  & 0.19  \\
0,1,2,3,4   & 95.9  & 0.19  \\
\bottomrule
\end{tabular}
\label{tab:layer}
\end{table}




\newpage

\end{document}